\documentclass[11pt]{article}
\usepackage{authblk}
\usepackage{coling2020}
\usepackage{times}
\usepackage{url}
\usepackage{latexsym}

\usepackage[colorinlistoftodos]{todonotes}

\usepackage{times}
\usepackage{soul}
\usepackage{url}
\usepackage[hidelinks]{hyperref}
\usepackage[utf8]{inputenc}
\usepackage[small]{caption}
\usepackage{graphicx}
\usepackage{amsmath}
\usepackage{amsthm}
\usepackage{booktabs}
\usepackage{algorithm}
\usepackage{algpseudocode}
\usepackage{amsmath}
\usepackage{amssymb}
\usepackage{multirow}
\usepackage{longtable}
\usepackage{comment}
\usepackage{caption} 
\usepackage{subcaption} 
\usepackage{csquotes}

\title{Generating Plausible Counterfactual Explanations for \\ Deep Transformers in Financial Text Classification}

\colingfinalcopy

\author[1]{\textbf{Linyi Yang}}
\author[1]{\textbf{Eoin M. Kenny}}
\author[2]{\textbf{Tin Lok James Ng}}
\author[3]{\textbf{Yi Yang}}
\author[1]{\textbf{Barry Smyth}}
\author[1]{\textbf{Ruihai Dong}}
\affil[1]{Insight Centre ,University College Dublin, Ireland}
\affil[1]{\tt {\{first.last\}@insight-centre.org}}
\affil[2]{University of Wollongong, Australia}
\affil[2]{\tt {jamesng@uow.edu.au}}
\affil[3]{The Hong Kong University of Science and Technology, Hong Kong, China}
\affil[3]{\tt {imyiyang@ust.hk}}

\begin{document}

\maketitle

\begin{abstract}
  Corporate mergers and acquisitions (M\&A) account for billions of dollars of investment globally every year, and offer an interesting and challenging domain for artificial intelligence. However, in these highly sensitive domains, it is crucial to not only have a highly robust and accurate model, but be able to generate useful explanations to garner a user's trust in the automated system. Regrettably, the recent research regarding eXplainable AI (XAI) in financial text classification has received little to no attention, and many current methods for generating textual-based explanations result in highly implausible explanations, which damage a user's trust in the system. To address these issues, this paper proposes a novel methodology for producing plausible counterfactual explanations, whilst exploring the regularization benefits of adversarial training on language models in the domain of FinTech. Exhaustive quantitative experiments demonstrate that not only does this approach improve the model accuracy when compared to the current state-of-the-art and human performance, but it also generates counterfactual explanations which are significantly more plausible based on human trials.
\end{abstract}

\section{Introduction and Related Work}
In recent years, large-scale, pre-trained transformer models have led to massive improvements on a wide range of natural language processing (NLP) tasks~\cite{devlin2018bert,liu2019roberta}, including financial technology applications~\cite{duan2018learning,yang2018explainable,xing2019sentiment,Yang20}. However, this impressive ability also coincides with an inherent lack of \textit{robustness} and \textit{transparency}, which undermines human trust in the prediction outcome. In the highly sensitive (and financially lucrative) area of FinTech, explainable financial text classification remains an open, and highly alluring question. To tackle this problem, this paper advances a novel approach which first applies robust transformer models (by leveraging adversarial training) on a real-world, up-to-date, self-collected mergers and acquisitions (M\&A) dataset, and then generating plausible, \textit{post-hoc}, counterfactual explanations. In the remainder of this section, we describe relevant work to both of these areas before detailing our contributions.

\subsection{Artificial Intelligence in Mergers and Acquisitions}
M\&As have reshaped the global business landscape for generations, and are having an accelerating impact on the world's economy as new technologies such as the internet, big data, and artificial intelligence disrupt many business sectors~\cite{yan2016modeling}. To appreciate this, a recent economic study provided strong evidence that M\&A deal rumours could influence the share price volatility of rumor target firms~\cite{ma2016investor}. In particular, they showed that, on average, M\&A rumors have a positive short term impact and a negative long term impact on the cumulative abnormal returns of the potential acquirers and targets. In the existing AI literature, focus here is typically on predicting likely M\&A targets~\cite{yan2016modeling}, and forecasting the likely success of M\&A~\cite{danbolt2016abnormal} for developing high-risk/high-reward investment strategies based on M\&A speculation~\cite{ji2009shrinking}. While the existing literature typically focuses on predicting likely M\&A acquirers and targets, in this work we address a distinct but related task: namely, whether a merger and acquisition \emph{rumor} is likely going to prove to be correct.

\subsection{Visualization-based Explanations}
To interpret a model's prediction, prior efforts have focused on either incorporating \textit{pre-hoc} analysis into the experimental design~\cite{Brunner2020On}, or developing \textit{post-hoc} analysis algorithms to select or modify particular instances of the dataset to explain the behavior of models~\cite{keane2020good,kenny2019twin}. Recent research~\cite{jain2019attention} shows that transformer models can not be perfectly explained from their intrinsic architecture, and a further work~\cite{Brunner2020On} provides strong evidence that self-attention distributions are not directly interpretable. For this reason, model-agnostic, \textit{post-hoc} explanation methods have come to the fore among these works for explaining text classification models, as they are easy to understand and do not require access to the data or the model~\cite{keane2020good}.

Towards \textit{post-hoc} explanation in NLP tasks, \cite{james2018beyond} proposes a popular way named contextual decomposition (CD) to quantify the importance of each individual word/phrase by computing the change to the model prediction when solely removing a word/phrase. Its hierarchical extensions \cite{singh2018hierarchical,Jin2020Towards} continue to refine the explanation algorithms that calculate and further visualize the individual phrase's importance. However, despite these visualization-based methods \cite{james2018beyond,singh2018hierarchical,Jin2020Towards} having achieved good results on a popular dataset of sentiment analysis (namely the Stanford Sentiment Treebank-2 [SST-2] dataset where human create the ground truth with their subjective judgement), how to generate explanations in more complex scenarios where human performance is worse than a model have not been well studied. As a result, the prior lines of visualization-based works cannot provide a clear boundary between positive and negative instances to human, whereas counterfactuals could provide \enquote{human-like} logic to show a modification to the input that makes a difference to the output classification~\cite{byrne2019counterfactuals}. Hence, \textit{post-hoc}, example-based explanation methods have received more and more attention in recent years~\cite{keane2020good}.

\subsection{Counterfactual Text Explanations}
Counterfactual explanations are renown for their explanatory ability in AI systems~\cite{wachter2017counterfactual}; specifically, they offer the ability to explain models (such as transformers) without having to \enquote{open the black-box}~\cite{grath2018interpretable}, by conveying causal information about what contributed to a given classification. To understand counterfactuals in the context of text classification, consider a sentiment classification task were a black-box model may classify \enquote{John loved the film} with a positive sentiment, and explain the prediction \textit{counterfactually} by presenting \enquote{John \textit{hated} the film}. Glossed, this latter text is the AI explaining the prediction by saying \enquote{f the word \textit{love} was replaced with the word \textit{hate}, I would have thought it was a negative sentiment}. This allows us to understand the main reasoning process behind the classifier in question, thus explaining the prediction causally. To understand the issue of counterfactual \textit{plausibility}, consider that the previous explanation may also generate a counterfactual which reads \enquote{John \textit{not} the film}. This text may \enquote{flip} the classification to the counterfactual class, but it is \textit{grammatically implausible}, and (arguably) very difficult to contextualize. The reason this is important is because humans avoid creating counterfactuals which are far from a \enquote{possible world}~\cite{wachter2017counterfactual}, and by extension wildly implausible~\cite{byrne2019counterfactuals,kenny2020generating}. In response to this, our work attempts to guarantee more grammatically plausible explanations, and does not rely on attention weights, nor is it constrained to a specific text domain.



\paragraph{Contributions and Paper Outline}
\begin{itemize}
    \item We present a novel dataset to the interesting and challenging problem of artificial intelligence in M\&A prediction.
    \item To the best of our knowledge, the present work is the first general approach to generate grammatically plausible counterfactual explanations for unstructured text classification.
    \item The primary technical contribution in this work is to generate grammatically \textit{plausible} counterfactuals by replacing the most important words with the antonyms (REP-SCD) based on pre-trained language models. Furthermore, two additional variants (removing/inserting works at the most important place, namely RM-SCD and INS-SCD) are proposed to guarantee counterfactual generations, albeit ones which are less plausible.
\end{itemize}

The remainder of this paper is organized as follows. Section 2 details our novel dataset and the pre-processing steps involved. Section 3 describes our adversarial training approach, with the sensitivity-based method for counterfactual explanation generation. Exhaustive experiments (both quantitative and human-based) show clear improvements in our method over current state-of-the-art, both in regards to classification accuracy, and explanation quality (see Sections 4 and 5). Finally, the implications of this work on XAI and future research is discussed.

\section{The Novel Mergers and Acquisitions Dataset}


\begin{table}[ht]
\centering
\begin{tabular}{lll}
\hline
Description & Number \\
\hline
\#Processed deal news total (2007-2019) & 4,098 \\
\#Train (2007-2014) & 3,120 \\
\#Validation (2015 - 2016) & 478 \\
\#Test (2017 - Aug 2019) & 500 \\
\#Unique companies and institutions & 1,406 \\\hline
\end{tabular}
\label{tab:plain}
\caption{The description of our dataset}
\end{table}

For this study we adopted a large-scale, up-to-date M\&A dataset collected from Zephyr, a comprehensive database of deal data from the \enquote{real world}. The dataset \footnote[1]{\url{https://www.bvdinfo.com/en-gb/our-products/data/specialist/zephyr}} contains 14,539 news articles or tweets on M\&A events between January 1st 2007, and August 12th 2019. Each instance corresponds to a specific editorial M\&A article which describes a possible deal between an acquirer and a target company (also including a few IPO rumours). Additionally, each datapoint also includes the deal outcome (see below), and the deal announcement data, if relevant. In this work, the deal outcome corresponds to the target class, and the raw dataset contains the following outcome types: \emph{complete} -- a deal between the acquirer and target companies concluded successfully; \emph{rumour} -- no deal materialized between the acquirer and target company; \emph{pending} -- a desired deal between the acquirer and target company has been confirmed, and at the time of data collection was deemed to be in-progress, but not yet complete; \emph{cancelled} -- a past potential deal between the acquirer and target companies has been confirmed, but it did not complete, and is no longer being pursued. 

In order to prepare the raw dataset for use in this study, a number of pre-processing steps were carried out:
\begin{enumerate}
    \item In this work we chose to focus on a binary classification task and, as such, removed instances with outcome types of \emph{cancelled} and \emph{pending}, leaving only those instances that correspond to \emph{completed} deals (the positive class) and \emph{rumours} (the negative class).
    \item We eliminated instances where \emph{both} acquiring and target companies were non-US, due to a tendency towards low-quality data; in other words, all of the instances in our dataset include a US Listed Company as either the acquirer or the target or both.
    \item Articles published within one day or after the deal announcement date were also removed, this is because our interest is in developing a prediction model that is capable of generating accurate predictions at least one day in advance of any deal outcome.
    \item Finally, the remaining instances are randomly over-sampled to ensure an even split between positive (completed) and negative (rumours) instances for each year.
\end{enumerate}

The result is a dataset of 4,098 instances (news articles and meta-data) which we split into training, validation, and testing sets on a year-by-year basis (see Table 1).

\section{Methodology}



The pipeline of our method is shown in Fig.~1. First, as a prerequisite, a transformer variant is fine-tuned on the M\&A prediction task, alongside adversarial training (which as we shall see is shown to be promising in this domain). Second, important words in the test instances are identified using a sampled contextual decomposition technique after the prediction. Third, a counterfactual explanation is generated by replacing these words with \textit{grammatically plausible} substitutes. As we shall see, although this method does not always guarantee a plausible counterfactual will be found, we propose two alternative methods which will, albeit with the possible trade-off of plausibility. These steps are detailed next.

\subsection{Step 1: Robust Transformer Classification Models}
As eluded to earlier, M\&A prediction is a highly sensitive domain, and despite adversarial training showing promise previously~\cite{goodfellow2014explaining,tsipras2018robustness}, it has never been tested in this domain. Hence, to try ensure a robust model which can simultaneously generate intelligible explanations, we explore its usage here compared to other popular approaches. Given a news article, we adopt the classical transformer architecture proposed by \cite{vaswani2017attention}. The original multi-head self-attention is subsequently applied to the \(k\)-th document \(\mathcal{D}^{(k)}\), which is calculated as follows:
\begin{equation}
\mathrm{MultiHead}= \text {Concat} \left(\mathrm{head}_{1}, \ldots, \mathrm{head}_{\mathrm{h}}\right) W^{O}
\end{equation}
\begin{equation}
\operatorname{head}_{\mathrm{j}}=\operatorname{Attention}\left({Q},{K},{V}\right)
\end{equation}

\begin{equation}
\begin{aligned}
Q = \mathcal{D}^{(k)} W_{j}^{Q}, K = \mathcal{D}^{(k)} W_{j}^{K}, V = \mathcal{D}^{(k)} W_{j}^{V}
\end{aligned}
\end{equation}%

\text{where} \(W_{j}^{Q}, W_{j}^{K}, W_{j}^{V} \in \mathbb{R}^ {{d} \times {d}} \) are weight metrics, and the attention is computed as:

\begin{equation}
\text { Attention }({Q},{K},{V})=\operatorname{softmax}\left(\frac{{QK}^{\top}}{\sqrt{d}}\right)\mathbf{V}
\end{equation}%
for input query, key and value matrices $Q, K, V \in \mathbb{R}^{n \times d}$. The \(h\) outputs from the attention calculations are concatenated and transformed using an output weight matrix \(W^{o} \in \mathbb{R}^{d h \times d}\).


\begin{figure*}[!t]
  \centering
  \includegraphics[width=\linewidth]{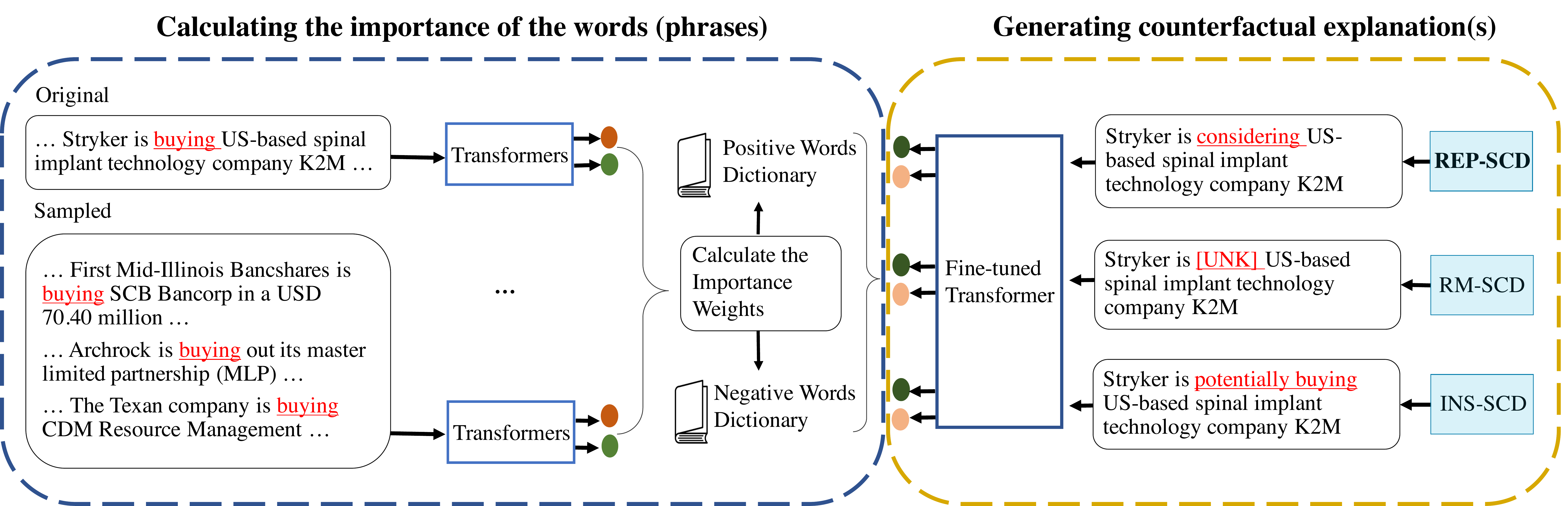}
  \caption{The pipeline of our methods, namely REP-SCD, RM-SCD, and INS-SCD. We show real examples of generating diverse counterfactual instances that flip the prediction result from \emph{completed} to \emph{rumour}. The original input has been changed by iteratively modifying words in order of their importance until the prediction matches the counterfactual class. The outputs (logits) of the predictions are represented in green, and orange points, respectively.}
\end{figure*} 

Additionally, the adversarial noise, treated as a form of regularization, is generated by the Fast Gradient Method (FGM) \cite{miyato2017adversarial} and Projected Gradient Descent (PGD) \cite{madry2018towards}. The idea of using adversarial perturbation is derived from the usage of adversarial attacks \cite{carlini2017towards} to evaluate the robustness of neural networks, while the recent advances of using the adversarial training in NLP models \cite{liu2020adversarial} inspires us to use it as a way of regularization. For each embedded word \(e\) in \(k\)-th news article \(\mathcal{D}^{(k)}\), the FGM computes its perturbation as follows:

\begin{equation}
{r_{fgm}=\epsilon \cdot g /\|g\|_{2}} \\ 
\text{ where }{g=\nabla_{e} L(\theta, (\mathcal{D}^{(k)}, y))}
\end{equation}

where  \(r_{fgm}\)  is the perturbation of \(e\), \(\theta\) denotes the current values of the parameters of the classifier, and \(L\) denotes the loss function (cross entropy) associated with the classifier. The perturbation can be easily computed using back-propagation. The projected gradient descent, which can be considered as a multi-step variant of the FGM, computes the perturbation of \(e\) iteratively:

\begin{equation}
    \begin{array}{l}{e_{t+1}=\Pi_{e+S}\left(e_{t}+\alpha g\left(\mathcal{D}^{(k)}_{t}\right) /\left\|g\left(\mathcal{D}^{(k)}_{t}\right)\right\|_{2}\right)} \\  {g\left(\mathcal{D}^{(k)}_{t}\right)=\nabla_{e} L\left(\theta, (\mathcal{D}^{(k)}_{t}, y)\right)}\end{array}
\end{equation}



where \(S=\left\{r \in \mathbb{R}^{d}:\|r\|_{2} \leq \epsilon\right\}\) is the constraint space of the perturbation, \(\Pi_{e+S}\) denotes the projection of a vector onto the feasible set \(e + S\), and \(\alpha\) is the step size. We use Adam optimizer with learning rate decay to train our model until convergence.


\begin{algorithm}[ht]
\caption{Plausible Counterfactual Instances Generation}
\textbf{Input:} Testing document example \(\mathcal{D}^{(k)}\)= \(\{{w}_{1},{w}_{2},...,{w}_{n}\}\), the corresponding ground truth label Y, pre-trained Mask Language Model MLM, negative pronouns list NP, fine-tuned transformer classifier C. \\
\textbf{Output:} Positive Word Dictionaries POS, Negative Word Dictionaries NEG, Plausible counterfactual example(s) \({D}^{(k)}_{cf}\)= \(\{{D}^{(k)}_{REP-SCD},{D}^{(k)}_{RM-SCD},...,{D}^{(k)}_{INS-SCD}\}\)
\begin{algorithmic}[1]
    \State Initialization: ${D}^{(k)}_{cf} \gets {D}^{(k)}$
    \For{each word $w_i$ in in \(\mathcal{D}^{(k)}\)}
        \If{the prev word $w_{i-1}$ is in NP}
            \State Creat the whole phrase $np_i$ by contextual decomposition
            \State Computer the importance score ${P}_{{w}_{i}} = -{P}_{{np}_{(i)}}$ via Eq.(7)
        \Else
            \State Computer the importance score ${P}_{{w}_{i}}$ via Eq.(7)
        \EndIf
    \EndFor \\
    Create dictionaries with words: $W_{POS}; W_{Neg}$, alongside the word positions ${pos}_{{w}_{i}}$ sorted by the descending order of their importance scores ${P}_{{w}_{i}}$.
    
    \For{each word position $pos_i$ in ${pos}_{{w}_{i}}$}
        \State $W_{Plausible} \gets MLM({D}^{(k)}_{mask\_{w}_{pos_i}})$, $W_{Plausible}^{'} \gets MLM({D}^{(k)}_{mask\_{w}_{pos_i\pm1}})$
        \If{${Y}^{(k)}$ == POS}
            \State $W_{Candidate}$, $W_{Candidate}^{'}$ $\gets$ Intersection ($W_{NEG}$ and $W_{Plausible}$), ($W_{NEG}$ and $W_{Plausible}^{'}$)
        \Else
            \State $W_{Candidate}$, $W_{Candidate}^{'}$ $\gets$ Intersection ($W_{POS}$ and $W_{Plausible}$), ($W_{POS}$ and $W_{Plausible}^{'}$)
        \EndIf
        \State ${D}^{(k)}_{rm} \gets {D}^{(k)}\_{w_{pos_i}} $
    \EndFor
    
    \For{each word $w_i$,$w_i^{'}$ in zip ($W_{Candidate}$,$W_{Candidate}^{'}$)}
        \State ${D}^{(k)}_{ins}$ $\gets$ Insert $w_i^{'}$ to ${D}^{(k)}_{mask\_{w}_{pos_i\pm1}}$
        \State ${D}^{(k)}_{rep}$ $\gets$ Replace $w_i$ with ${D}^{(k)}_{mask\_{w}_{pos_i}}$
        \If{$C({D}^{(k)}_{rm} , {D}^{(k)}_{ins} , {D}^{(k)}_{rep}) \neq$ Y}
            \State Add ${D}^{(k)}_{rm} , {D}^{(k)}_{ins} , {D}^{(k)}_{rep}$ to the set ${D}^{(k)}_{cf}$
        \EndIf
    \EndFor
    \State \Return{${D}^{(k)}_{cf}$}
\end{algorithmic}
\end{algorithm}

\subsection{Step 2: Context-Independent Word Importance}
To calculate the context independent importance up to one word, we adopt the \textit{sensitivity of contextual decomposition} technique from \cite{madry2018towards} which removed part of inputs from the sequence text to evaluate a model's sensitivity to them, thereby allowing for the identification of important features. In its hierarchical extensions -- Sampling and Contextual Decomposition (SCD), \cite{Jin2020Towards} mask out the phrase \(p\) from the input while the max sequence length \(N\) is set to 40. However, the average input length in our data is much larger than 40. We, therefore, propose a phrase-level removing method only if the phrase starts with the negative pronouns or limitations. Otherwise, only a single word will be removed. For example, in the sentence \enquote{the deal is not closing currently}, the attribution of \enquote{closing} should be positive while the attribution of \enquote{not closing} should be negative. In this situation, we remove the whole phrase \enquote{not closing} together to calculate the influence in terms of the logits change in the output layer of the transformer and then assign the negative score to the word \enquote{closing}. 

Given a phrase \(p\) starting with the negative limitations in the \(k\)-th document $\mathcal{D}^{(k)}$, we sample the documents which contain the same phrase \(p\) to alleviate the influence by chance when there are multiple shreds of evidence saturating the prediction. For example, in the source \enquote{JPMorgan is closing in on a deal, sources close to the situation are optimistic for deal completion}, if we only remove the word \enquote{closing}, the prediction would not be changed so much. In this sampling way, the proposed context-independent importance of word and phrase is more robust to saturation. The formula for calculating the importance can be written as:
\begin{equation}
\phi(\mathbf{p}, \widehat{\mathcal{D}^{(k)}})=\mathbb{E}_{\widehat{\mathcal{D}^{(\beta)}}}\left[l\left(\widehat{\mathcal{D}^{(\beta)}}; \widehat{\mathcal{D}}\right)-l\left(\widehat{\mathcal{D}^{(\beta)}} \backslash \mathbf{p} ; \widehat{\mathcal{D}}\right)\right]
\end{equation}

where \(\mathcal{D}^{(\beta)}\) denotes the resulting document after masking out a single token or a phrase starting with the negative pronoun in the length of \(N\) surrounding the phrase \(\mathbf{p}\). we use \(l\left(\widehat{\mathcal{D}^{(\beta)}} \backslash \mathbf{p} ; \widehat{\mathcal{D}}\right)\) to represent the model prediction logits after replacing the masked-out context. \(\backslash \mathbf{p}\) indicates the operation of masking out the phrase \(p\) in a input file sampling from the testing set \(\mathcal{D}\). 

As an aside, the resulting top 15 most influenced words are shown in Table 2. In total, there are 123 positive words and 155 negative words in the dictionaries. We can see the average influence score of positive words (0.637) is higher than the negative words (0.385). It may reveal that positive words usually contain more powerful clues in predicting the M\&A deal. That would be interesting to see which kind of words in the sources illustrate the deal is more likely to be completed in the future and which kind of words would be likely to kill the deal.

\begin{table}[]
\centering
\begin{tabular}{llll}
\hline
Positive Words & Sensitivity & Negative Words & Sensitivity \\ \hline
announced & 5.841 & talks & 4.674 \\
line & 5.715 & could & 2.484 \\
announcement & 4.469 & flag & 2.236 \\
agreement & 3.378 & diligence & 1.363 \\
acquiring & 3.342 & considering & 1.196 \\
completion & 2.727 & time & 1.186 \\
agreed & 2.429 & may & 1.085 \\
closing & 2.125 & looking & 0.983 \\
consideration & 1.994 & this & 0.972 \\
prevailed & 1.639 & when & 0.914 \\
acquire & 1.520 & potentially & 0.870 \\
paid & 1.461 & if & 0.847 \\
disclosed & 1.403 & intention & 0.836 \\
selling & 1.385 & year & 0.812 \\
could & 1.360 & takeover & 0.790 \\ \hline
\end{tabular}
\label{tab:plain}
\caption{Top 15 most influenced words towards the M\&A prediction. The influence score for each word is calculated and added up by Sampling and Contextual Decomposition (SCD) on the testing set.}
\end{table}

\subsection{Step 3: Counterfactual Instance Generation}
As shown in Algorithm 1, we summarize three different counterfactual generation methods, namely, the primary technique which generates grammatically plausible counterfactuals (REP-SCD), and two further variants to guarantee counterfactual generation (RM-SCD and INS-SCD). We combine these three methods to alleviate a major issue in counterfactual explanation, that is, there is no guarantee that for a given example a counterfactual instance is found. Our main technique identifies the most important word(s) in a test instance using SCD and replaces them with the intersection of grammatically plausible substitutes [using masked language model (MLM)] and words in the reverse emotional dictionary. The raw document content \(\mathcal{D}^{(k)}\) itself is taken as input, and MLM outputs \(p(\cdot | \mathcal{D}^{(k)})\) for each masked position. After all masked positions are infilled, we get the reconstructed document:
\begin{equation}
    \widehat{\mathcal{D}^{(k)}}=\mathrm{MLM}(\mathcal{D}^{(k)}).
\end{equation}
We iterative repeat this operation at the most important word positions ranked by SCD until the reconstructed document ultimately moves the model's classification towards the opposing class. Notably, there may be more than one counterfactual explanation corresponding with the original text instance.


\section{Experiment 1: Financial Text Classification with Robust Transformers}
In this section we describe the results of a comprehensive evaluation of classification accuracy, comparing a variety of different classification baselines (including a human baseline) to our adversarial transformer approach.

\subsection{Methods Used}
The baselines used can be grouped into several distinct categories: human evaluations -- traditional machine learning approaches (SVM) -- classical deep learning approaches (CNN \cite{kim2014convolutional}, BiGRU \cite{bahdanau2014neural} , and HAN \cite{yang2016hierarchical}) -- and various transformer approaches with/without pruning strategies. These transformer-based models are generally considered to provide the current state-of-the-art in text classification. We reproduce these baselines based on the Transformers.\footnote[2]{\url{https://github.com/huggingface/pytorch-transformers}} 

\paragraph{Acquiring a human baseline}
As a baseline, we asked 26 participants which were experts in economics and finance to predict M\&A events by completing 50 M\&A evaluation questionnaires. The participants consisted of Ph.D. students, and academics from the fields of economics/finance. All participants were either native English speakers or had a high degree of English competence. Each questionnaire provided information on ten M\&A cases/instances, sampled randomly without replacement from the test set. In addition, the news articles available in the dataset that were published before the deal announcement were also provided. The questionnaire asked the participant to predict the outcome of the deal (complete or rumour), and to state their confidence in this prediction.

\subsection{Classification Results}
In line with best practice, model hyper-parameters are tuned using the validation set. In particular, the maximum sequence length is set as 256, and the size of transformers are all set as large. All experiments are using the conventional Matthews Correlation Coefficient \emph{(MCC)}, accuracy and \emph{F1} metrics. The classification results are summarized in Table 3 with \emph{Random Guess} used to provide a lower-baseline based on chance. While the human evaluators performed better than chance their ability to predict deal outcomes is limited when compared to the more sophisticated machine models that follow. These results are particularly compelling as the human evaluators had considerable domain expertize.

Each of the machine learning approaches offer substantial improvements over the human evaluators and a clear separation can be seen between traditional machine learning (with MCC scores in low 0.7 range/F1 scores in the low 0.8 range), classical deep learners (with  MCC scores in the range 0.73-0.74/F1 scores in the range 0.84-0.85), and recent transformer-based models (MCC$>$0.75/F1$>$0.87).

We further evaluate the relative influence of the adversarial perturbation to test the robustness of the models. We find that all variants of the transformer \cite{lan2019albert,sanh2019distilbert} benefit from the adversarial perturbation during the training process in terms of the prediction results in the practice. For exploring the reason why the optimal transformer classifier can outperform the human test a lot -- 39\%, we take the best performed model -- RoBERTa \cite{liu2019roberta} with adversarial training as our optimal classifier in the following experiments for generating the plausible counterfactual explanations.  

\begin{table}[]
\centering
\begin{tabular}{llll|llll}
\hline
Evaluation & MCC & Accuracy & F1 & Evaluation & MCC & Accuracy & F1\\ 
\hline
\textbf{Baselines} & & & & \textbf{Transformers} &  &  & \\
Random Guess & 0.013 & 0.510 & 0.462 & ALBERT & 0.768 & 0.882 & 0.879\\
Human Evaluation & 0.307 & 0.640 & 0.672 & +Ad. & 0.780 & 0.890 & 0.888 \\
\textbf{Traditional ML} &  &  &  & DistilBERT & 0.750 & 0.874 & 0.877  \\
SVM(TF-IDF) & 0.701 & 0.816 & 0.816  & +Ad. & 0.784 & 0.890 & 0.891  \\
\textbf{Classical DL} &  &  &  & BERT-WWM & 0.751 & 0.874 & 0.879 \\
CNN-Text & 0.729 & 0.848 & 0.847 & +Ad. & \textit{\textbf{0.788}} & \textit{\textbf{0.894}} & 0.894 \\
BiGRU & 0.734 & 0.836 & 0.849 & RoBERTa & 0.780 & 0.892 & 0.888  \\
HAN & 0.742 & 0.848 & 0.853 & +Ad. & \textit{\textbf{0.788}} & \textit{\textbf{0.894}} & \textit{\textbf{0.895}}  \\
\hline
\end{tabular}
\label{tab:plain}
\caption{Evaluations performed by human, machine learning, deep learning, and transformer-based models, alongside the ablation study for adversarial training (indicate as +Ad.). The scores in bold and italics indicate the best performance across all approaches.}
\end{table}

\section{Experiment 2: Generating Plausible Counterfactual Explanations}
Interpretability is an increasingly important property for many deep learning techniques, including computer vision and natural language processing~\cite{kenny2019twin}, especially in critical tasks such as financial text classification; high-value investment decisions demand a reasonable level of interpretability if investors are to trust the predictions that come for a system such as the one described in this work. In this section, we describe the qualitative analysis for each of our methods. Subsequently, we show the evaluation of user studies compared to the existing example-based explanation methods.

\subsection{Qualitative Analysis for the Resulting Counterfactual Instances}
In qualitative analysis, we identified five typical patterns among the generated counterfactual instances as shown in Table 4 where we highlight the changing parts. Based on the 500 testing examples, we guarantee that there is at least one counterfactual instance corresponding with the original input. We gain insight into which aspects are causally relevant by comparing the original context to the revised context which can flip the classifier's prediction.

\newcommand\boldblue[1]{\textcolor{blue}{\textbf{#1}}}
\newcommand\boldred[1]{\textcolor{red}{\textbf{#1}}}
\newcommand\boldorange[1]{\textcolor{orange}{\textbf{#1}}}
\newcommand\boldpurple[1]{\textcolor{purple}{\textbf{#1}}}

\begin{table*}[ht]
\begin{center}
\begin{tabular}{p{0.35\textwidth}p{0.6\textwidth}}

\hline
{ \textbf{Types of Algorithms}} & { \textbf{Examples}} \\ \hline
{ } & { Ori: Professional vacation services provider ILG is \boldblue{considering} a merger with Diamond Resorts International...} \\
\multirow{-2}{*}{{ \begin{tabular}[c]{@{}l@{}}REP-SCD:\\Replacing with the certainty word\end{tabular}}} & { Rev: Professional vacation services provider ILG is \boldblue{announcing} a merger with Diamond Resorts International...} \\ \hline
{ } & { Ori: Vivendi is in early discussions to sell a 10.0 per cent stake in Universal Music Group (UMG) to Tencent for roughly EUR 3.00 \boldblue{billion}...} \\
\multirow{-2}{*}{{ \begin{tabular}[c]{@{}l@{}}REP-SCD:\\Changing the deal value\end{tabular}}} & { Rev: Vivendi is in early discussions to sell a 10.0 per cent stake in Universal Music Group (UMG) to Tencent for roughly EUR 3.00 \boldblue{million}} \\ \hline
{ } & { Ori: Stryker \boldred{is buying} US-based spinal implant technology company K2M Group Holdings for USD 1.40 billion in cash} \\
\multirow{-2}{*}{{ \begin{tabular}[c]{@{}l@{}}INS-SCD:\\Recasting \(fact\) as \(hoped\) \(for\)\end{tabular}}} & { Rev: Stryker \boldred{is potentially buying} US-based spinal implant technology company K2M Group Holdings for USD 1.40 billion in cash} \\ \hline
{ } & { Ori: WPP has \boldred{confirmed} the recent speculation that it has entered into exclusive negotiations with private equity firm Bain Capital...} \\
\multirow{-2}{*}{{\begin{tabular}[c]{@{}l@{}}INS-SCD:\\ Inserting the negative word\end{tabular}}} & Rev: WPP has \boldred{not confirmed} the recent speculation that it has entered into exclusive negotiations with private equity firm Bain Capital...\\ \hline
{ } & { Ori: This suitor is the Namdar and Washington Prime consortium, the insiders noted, adding that there can be \boldorange{no certainty} a deal will complete...} \\
\multirow{-2}{*}{{ \begin{tabular}[c]{@{}l@{}}RM-SCD:\\Removing the negative limitation(s)\end{tabular}}} & { Rev: This suitor is the Namdar and Washington Prime consortium, the insiders noted, adding that there can be \boldorange{certainty} a deal will complete...} \\ \hline
\end{tabular}
\end{center}
\label{tab:plain}
\caption{Most prominent categories of counterfactual explanations generated by our algorithms, namely RM-SCD, REP-SCD, and INS-SCD for M\&A Predictions. Ori and Rev are short for original and revised instances, respectively.}
\end{table*}


\subsection{Human Evaluation for the Explanation}
We implement interpretation experiments on the optimal fine-tuned transformer classifier. While an explainable model trained with supervised learning is a common method to interpret the results of text classification \cite{wallace2019allennlp}, the self-supervised learning explainable frameworks have been scarcely found. Meanwhile, the work in \cite{kaushik2019learning} consider similar types of edits to generate counterfactually-revised data, however, all of the instances are generated by human which greatly limits the expansibility of the method. To comprehensively evaluate the performance of our method, we consider a state-of-the-art example-based explanation framework for comparison, namely HotFlip~ \cite{ebrahimi2017hotflip}, which uses gradients to identify important words and then flip it with the adversarial word which can cause the maximum change in gradients.

For user evaluation, here we ask domain experts in finance to rate our explanations on two aspects, (1) how \textit{plausible} (mainly in terms of grammar and comprehension) it is, and (2) how \textit{reasonable} it is (i.e., does the explanation make sense). We compare our method to Hotflip - the current state-of-the-art framework for counterfactual explanation - at the time of writing. Each score is measured on a scale of 1-5, where 5 is the best, and 1 is the worst. We randomly sample 100 examples from the testing set for 5 participants to answer (20 examples per person). By combining the REP-SCD, RM-SCD, INS-SCD together, our method achieves significantly higher ranking score compared to HotFlip, more specifically, 2.35 score improvements (4.35/2.00) were made regarding plausibility while 0.85 score improvements (4.00/3.15) were made on reasonableness, showing a $p$-value less than 0.001 and 0.05, respectively. Hence, there is compelling evidence that our method can generate counterfactual explanations which are more \textit{plausible} and \textit{reasonable}.

\section{Conclusion and Future Work}
In this work, we pursued a new research problem of M\&A prediction. Our transformer-based classifier leveraged the regularization benefits of adversarial training to enhance model robustness. More importantly, we built upon previous techniques to quantify the importance of words and help guarantee the generation of plausible counterfactual explanations with a masked language model in financial text classification. The results demonstrate superior accuracy and explanatory performance compared to state-of-the-art techniques. An obvious extension would be to include canceled deals into the classifier, or to predict novel M\&A events based on market descriptions of companies (e.g., scale, finances, and target markets). Moreover, additional financial events (e.g., misstatement detection and earnings call analysis) is yet another related task to be considered for further research. 

\section*{Acknowledgment}
We would like to thank Tianhao Fu, Yimeng Li, Yang Xu and Prof. Mark Keane for their helpful advice and discussion during this work. Also, we would like to thank the anonymous reviewers for their insightful comments and suggestions to help improve the paper. This research was supported by Science Foundation Ireland (SFI) under Grant Number $SFI/12/RC/2289$\_$2$. 




\bibliographystyle{coling}
\bibliography{coling2020}

\end{document}